\documentclass[a4paper]{article}

\usepackage[psfloats]{titleps}
\newpagestyle{main}{%
\setfoot{Accepted to Interspeech 2021}{}{}
}

\usepackage{INTERSPEECH2021}

\newcommand{\etal}{\textit{et al.~}}

\title{Fast Text-Only Domain Adaptation of RNN-Transducer Prediction Network}
\name{Janne Pylkkönen$^1$, Antti Ukkonen$^{1,2}$, Juho Kilpikoski$^1$, Samu Tamminen$^1$, Hannes Heikinheimo$^1$}

\address{$^1$Speechly, Finland\\$^2$Department of Computer Science, University of Helsinki, Finland}
\email{firstname@speechly.com}

\begin{document}

\maketitle
\begin{abstract}
Adaption of end-to-end speech recognition systems to new tasks is
known to be challenging. A number of solutions have been proposed
which apply external language models with various fusion methods, possibly
with a combination of two-pass decoding. Also TTS systems have been used to
generate adaptation data for the end-to-end models. In this paper we show that
RNN-transducer models can be effectively adapted to new domains using only small
amounts of textual data. By taking advantage of model's inherent structure,
where the prediction network is interpreted as a language model, we can apply
fast adaptation to the model. Adapting the model avoids the need for complicated
decoding time fusions and external language models. Using appropriate
regularization, the prediction network can be adapted to new domains while still
retaining good generalization capabilities.
We show with multiple ASR evaluation tasks how this method can provide
relative gains of 10--45\% in target task WER. We also share insights
how RNN-transducer prediction network performs as a language model.
\end{abstract}

\thispagestyle{main}  

\section{Introduction}

Over the recent years, the focus in automatic speech recognition research has shifted
from hybrid models to end-to-end (E2E) systems.
Traditional hybrid models consist of separate models for acoustic,
language, and pronunciation \cite{Hinton12, Zhang20}, whereas
E2E models integrate all of these into a single network
\cite{Graves12, Graves13, Chan16, He19}. The benefit of the hybrid models is that they can take
advantage of different data sources, especially large amounts of text-only data.
End-to-end models, on the other hand, are trained from matched speech and transcriptions,
so their exposure to different language content is more limited.

A particularly interesting E2E architecture is
the RNN-transducer (RNN-T) \cite{Graves12, Graves13}, which provides
state-of-the-art performance in a wide variety of streaming applications \cite{He19, Li20}.
Despite being an E2E architecture, RNN-T lends itself for a compelling
interpretation as having separate language and acoustic models.
However, some recent research have concluded that such an interpretation may not always
hold well \cite{Ghodsi20}. Even though it is possible to initialize the RNN-T
prediction network from a large text corpus, it has been unclear how much of
the predictive power of the initial language model (LM) remains
after the RNN-T has been trained with speech data.

To customize the E2E models for a particular domain, several methods have been
proposed \cite{Kannan18, He19, Shan19, Li20}, including application of external LMs,
and using TTS-generated data to fine-tune the network. Fusion methods require changes
to the model and/or decoding, whereas TTS-adaptation is a straightforward extension
of model fine-tuning. One of the most applied adaptation methods is the shallow
fusion \cite{Kannan18, He19,  Meng21}, where external language model scores are added
to the RNN-T scores during decoding.

\begin{figure}[t]
  \centering
  \includegraphics[width=0.85\linewidth]{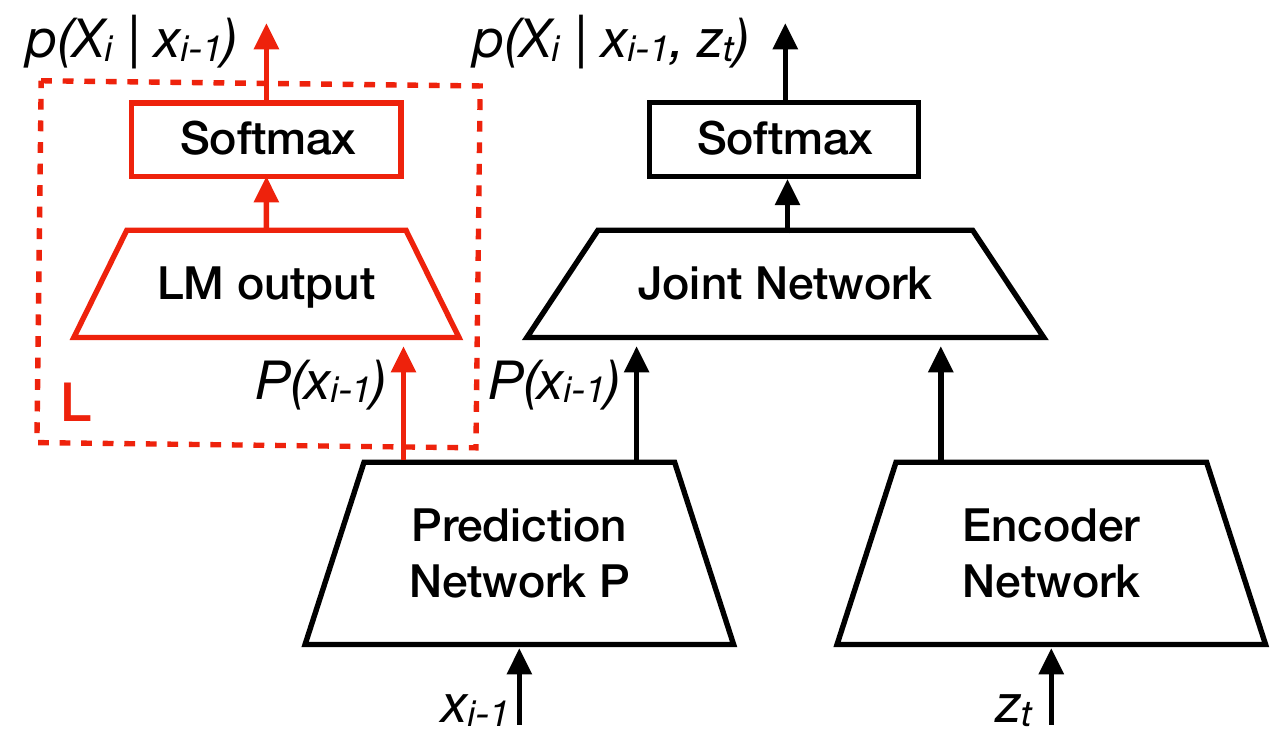}
  \caption{We propose a domain adaptation method for fine-tuning the prediction network $P$ of a trained RNN-T (shown in black). We first train a temporary LM output component denoted $L$ (shown in red) that enables the fine-tuning of $P$ with generic neural LM adaptation methods. Neither $L$ nor other changes to the RNN-T model are required when decoding.}
  \label{fig:rnnt-architecture}
\end{figure}

In this paper
we present a simple yet effective RNN-T adaptation method, which requires only
textual data. No speech data is involved in the adaptation process, not even via
a TTS system. This text-only adaptation can be performed quickly, and it does
not require any modifications to the decoding or model inference.
The only additional requirement is the estimation of
a temporary LM output layer on top of the prediction network (Fig.~\ref{fig:rnnt-architecture}).
With this output layer and suitable regularization,
the prediction network can be adapted as a neural LM,
while still remaining as a part of the RNN-T network.
Using this RNN-T adaptation
we observe 10--45\% relative word error rate (WER) improvements in target tasks.
In contrast to shallow fusion,
we show that these benefits can be obtained
without substantial degradation in out-of-adaptation-domain accuracy.

Besides their practical applicability,
our results also contribute to recent discussion about
the role of the RNN-T prediction network \cite{Ghodsi20, Meng21}.
In particular, we demonstrate that it is useful to
interpret the RNN-T prediction network as having characteristics of an LM.
This opens possibilities to using algorithms with E2E architectures
that have previously only been applicable to hybrid models.

\section{Adaptation of RNN-T Prediction Network}

\subsection{RNN-T architecture}
The RNN-transducer, shown in black in Fig.~\ref{fig:rnnt-architecture}, was first proposed by Graves \cite{Graves12, Graves13}, and later refined by others \cite{Rao17, He19}.
In RNN-T models both the prediction and encoder components are recurrent neural networks,
typically LSTM stacks. Compared to simpler encoder networks trained with connectionist temporal
classification (CTC) criterion \cite{Graves06, Hannun14}, RNN-T introduces a separate prediction network
to condition the prediction of the next token $x_i$ to the past emissions $x_1, \ldots x_{i-1}$
in addition to the acoustic input $z_t$.

RNN-T is attractive especially when streaming decoding is required, as the decoding uses only the left
context for predicting the next token. Streaming decoding requires that the recurrent hidden layers are
implemented as uni-directional layers.

\subsection{Algorithm overview}
Our motivation for adapting a fully-trained RNN-T model is to quickly customize the network
to different target tasks
with text-only data.
Moreover, we seek to do this
without introducing further components to the model, and
without increasing decoding complexity.
In search of such a fast text-only domain adaptation method,
we view the RNN-T prediction network as a neural language model.

Let $P$ denote the prediction network,
$D_t$ the transcriptions used to train the input RNN-T,
and $D_a$ the in-domain adaptation texts.
At the core of our approach is to
fine-tune the parameters of $P$ with $D_a$,
just as one would adapt a neural LM in general.
However, as part of an RNN-T model,
the output of $P$, denoted $P(x)$, is
an internal feature representation
rather than a conditional probability distribution
over the $i$th token given the previous tokens,
denoted $p(X_i \mid x_{i-1})$,
as required from an LM.
Simple neural LM adaptation techniques
cannot thus be applied directly to $P$, nor the RNN-T.
We propose a two-step solution to this problem:
\begin{enumerate}
\item {\bf Pre-processing step}
With transcriptions in $D_t$,
while keeping $P$ fixed,
train a new LM output component to $P$ (red in Fig.~\ref{fig:rnnt-architecture}) that
outputs the conditional distribution $p(X_i \mid x_{i-1})$.
Let $L$ denote this LM output component.
\item {\bf Adaptation-step}
With transcriptions in $D_a$,
fine-tune the LM induced by $P$ and $L$,
but keep $L$ fixed.
Output an RNN-T where $P$ replaces the
original prediction network.
\end{enumerate}

Note that the composition of $P$ and $L$ is a generic neural LM, and
hence in the {\em pre-processing step}
$L$ can be trained with text input using a
standard autoregressive LM loss, such as cross-entropy.
The training of $L$ is done over the transcriptions in $D_t$,
to avoid
introducing a mismatch with $P$ that was originally trained with those same
transcriptions. With $L$ trained,
we can proceed to the {\em adaptation step} in which
we use the texts in $D_a$ to
fine-tune the LM formed by $P$ and $L$.

Crucial to our approach is that when training $L$,
the parameters of $P$ are fixed,
and conversely, when fine-tuning $P$,
the parameters of $L$ are frozen.
This forces $L$ to first learn the same distribution
of $P(x)$ that the RNN-T joint network expects,
and then makes $L$ act as a regularizing constraint, preventing
$P(x)$ from becoming incompatible with the joint network.

Note that $L$ is required only when fine-tuning $P$,
it is {\em not} used during decoding.
Overall, decoding is unaffected by this adaptation procedure,
as no changes are introduced to the RNN-T architecture.
As the LM output component $L$ we chose
to use a single feed-forward layer followed by a softmax,
similar to the RNN-T joint network.
In both steps, standard cross-entropy loss function was used for optimization.

\subsection{Regularization and optimization details}

We want to further ensure that $P$ does not change in ways that would be harmful for RNN-T decoding.
To balance the fit to adaptation data $D_a$ and
the transcriptions $D_t$ used during RNN-T training,
we propose to augment the cross-entropy loss used in the adaptation step with a term that
penalizes changes in the predictions observed with common utterances.
This additional
regularization should promote the generalization ability of $P$.

To formalize this,
denote the original, non-adapted prediction network with $P^*$,
and let $x = (x_0, \ldots, x_n)$ denote an utterance with $n$ tokens,
prepended with a start token $x_0$.
Let
$p(X_i \mid x_{i-1}) = L(P(x_{i-1}))$ and
$p^*(X_i \mid x_{i-1}) = L(P^*(x_{i-1}))$ denote
the distributions output by the LMs
formed by $P$ and $L$ and $P^*$ and $L$, respectively,
for input token $x_i$.
The balancing loss term $\ell_b$ for input $x$ is then defined as
\begin{equation}
\label{eq:kld}
    \ell_b(x, P) = \frac{1}{n} \sum_{i=1}^n \mathrm{KLD}\left(p(X_i \mid x_{i-1}), p^*(X_i \mid x_{i-1})\right),
\end{equation}
where $\mathrm{KLD}$ is the Kullback-Leibler divergence.
The term $\ell_b(x, P)$ thus measures the difference between the
next-token distributions induced by
$P$ and $P^*$ for an utterance $x$.

To prevent $P$ from over-fitting to the adaptation corpus $D_a$,
we use Eq.\ref{eq:kld} together with a set of utterances
that are not part of $D_a$.
We generate these utterances by sampling the LM induced by $P^*$.
For each adaptation example $x$ in $D_a$,
another utterance $\hat{x}$ of similar length is generated.
Let $D_b$ denote the set of these utterances.
We also introduce a second regularization term
which penalizes the
drifting of the weights of $P$ from
their original values in $P^*$,
defined as $\ell_n(P) = \|P - P^*\|_2$.
The final adaptation loss function is hence:
\begin{equation}
\label{eq:loss}
\ell(P) = \sum_{x \in D_a} \frac{\mathrm{CE}(x, P)}{|D_a|} \; +
          \frac{w_b}{|D_b|} \sum_{x \in D_b} \ell_b(x, P) \; +
          w_n \ell_n(P),
\end{equation}
where $\mathrm{CE}(x, P)$ is the standard cross-entropy loss of
the LM induced by $P$ for utterance $x$, and
$w_b$ and $w_n$ are weights of the balancing and norm losses, respectively.

\section{Experiments}

\subsection{ASR system}

Our RNN-T architecture is similar to one presented by He \etal \cite{He19},
utilizing layer-normalized LSTMs with projection layers.
The first layer of the encoder network is a convolutional layer with 1536 filters. As
input features we use 32 dimensional MEL filterbank energies, emitted 100 times a second. The convolutional layer reduces the frame rate to 30ms/frame for the first two LSTM layers, after which the time reduction layer halves the frame rate to 60ms/frame. There are 7 LSTM layers in total, each with 1536 memory cells, and a projection dimension of 640. For RNN-T training,
the encoder network was initialized by training it first with a CTC loss function.

The prediction network consists of 2 layers of layer-normalized LSTMs with projection layers, both with 1536 cells and a projection dimension of 640. It was initialized as a neural
LM of the same architecture, augmented with a softmax output layer, and trained
over a 20G-word subset of the English Oscar corpus \cite{Suarez19}. During the initialization,
the same word piece lexicon was used as with the joint network.

The joint network is a simple feed-forward network, which takes the inputs from the projection layers of the prediction and encoder networks. Softmax activation is applied to the joint network output, resulting in a 1001-dimensional output vector. The output encodes 1000 word pieces and a blank symbol.

The training of the network was done with the RNN-T loss function until no error reduction over the training-time
development set was observed. The network training was done with SGD using a slowly decaying learning rate.
To reduce overfitting to the training data we applied SpecAugment \cite{Park19} throughout the training.

During inference the RNN-T model was used with a beam-search decoder,
which restricts the maximum number of expanded hypotheses from frame-to-frame.
For the experiments in this paper, this limit, the beam width, was set to 5.
For shallow fusion experiments, we converted
an n-gram model into an FST, and added weighted LM scores to the RNN-T scores
after each emission of a word piece.

\subsection{Data}

The RNN-T model was trained using three public English speech corpora: LibriSpeech \cite{Panayotov15},
English Common Voice \cite{Ardila20} (version 5.1, June 2020), and Ted-lium 3 \cite{Hernandez18}.
We chose to use only utterances with durations in the 1 -- 17s range, resulting in 1.57M,
with total of about 2770h of audio.
During the training, a subset of Common Voice development set was used to monitor
the progress and determine when to stop.
The training ran about 20 epochs over the data.

The amount of speech training data is modest for an E2E model, and does not result in state-of-the-art results. However, we chose to use only well-known publicly available corpora to enable experiment reproducibility. We feel that the experiments remain informative
and representative to the possible gains achievable by the proposed adaptation method. However,
we decided to publish adaptation results also with our production model, which shares a similar
RNN-T architecture, but has been trained with an order of magnitude more data.

The evaluations were carried out with three different datasets: Ted-lium 3 \cite{Hernandez18}, ATIS3 \cite{Dahl1994}, and Slurp \cite{Bastianelli20}. For ATIS3 the evaluation used the December 1993 test set with the Crown microphone audios used where multiple audios were available. Adaptation used all unique transcriptions from the complete ATIS3 training corpus. For Slurp the evaluation used the headset audio of each speaker for all test set transcriptions and for adaptation all the unique transcriptions from the non-synthetic training set. The Slurp dataset is divided to 18 distinct real-life application scenarios, and the evaluations were carried out both using per-scenario subsets and using the scenarios pooled into one dataset. Table \ref{tab:eval_data} summarizes the different evaluation sets.
In all the experiments, the ASR accuracies were measured with word error rate.

\begin{table}[t]
  \caption{Amount of adaptation (text) and evaluation (audio) utterances, and the size of the vocabulary in the datasets used in the experiments.}
  \label{tab:eval_data}
  \centering
  \begin{tabular}{ lccc }
    \toprule
    \textbf{Dataset} & \#Adaptation utts & \#Eval utts & Vocabulary \\
    \midrule
    ATIS3  & 6355 & 965 & 1080 \\
    Slurp & 10680 & 4173 & 5168 \\
    Ted-lium & -- & 1155 & 3652\\
    \bottomrule
  \end{tabular}
\end{table}

The weights for the balancing and norm loss were optimized over the Slurp development set, where the model adaptation was carried out over the Slurp training transcriptions. This optimization showed that the adaptation method is not particularly sensitive to the exact values of the loss weights.
All the experiments (except those in Section \ref{sec:generalization}) were then carried out with the same weights: $w_b = 0.8$, $w_n = 0.05$. The adaptation was stopped when the prediction network $L_2$-norm change exceeded value $4$.

\subsection{The effect of adaptation to ASR accuracy}

\begin{table}[t]
  \caption{WER for unadapted and adapted models over evaluation corpora, and relative WER reduction.}
  \label{tab:adapted_wer}
  \centering
  \begin{tabular}{ lccc }
    \toprule
    \textbf{Dataset} & Unadapted & Adapted & WERR-\% \\
    \midrule
    ATIS3  & 15.9\% & 11.9\% & -25.2\% \\
    Slurp (pooled) & 42.8\% & 38.6\% & -9.8\% \\
    Slurp (scenario) & 42.8\% & 37.3\% & -12.9\% \\
    \midrule
    ATIS3 (prod)  & 9.7\% & 5.4\% & -44.7\% \\
    Slurp (scen; prod) & 27.4\% & 23.4\% & -14.6\% \\
    \bottomrule
  \end{tabular}
\end{table}

Table \ref{tab:adapted_wer} shows the word error rates of unadapted and adapted
models over two different datasets.
For ATIS3 and Slurp evaluations, the models were adapted with dedicated held-out transcriptions.
For Slurp, both the result for the pooled dataset and
the overall result of per-scenario adaptations are shown.
The results are shown with the experiment model described above, as well as our production model.
With the experiment model, the adaptation ran for 16 and 23 epochs
over the ATIS3 and pooled Slurp adaptation data,
respectively, until the maximum norm change condition was reached.
The adaptation provides significant improvements in the accuracy, and
it is achieved with remarkably small adaptation sets:
only 6355 utterances for ATIS3, and 10680 utterances for Slurp.

\subsection{Generalization of the adapted models}
\label{sec:generalization}

\begin{figure}[t]
  \centering
  \includegraphics[width=\linewidth]{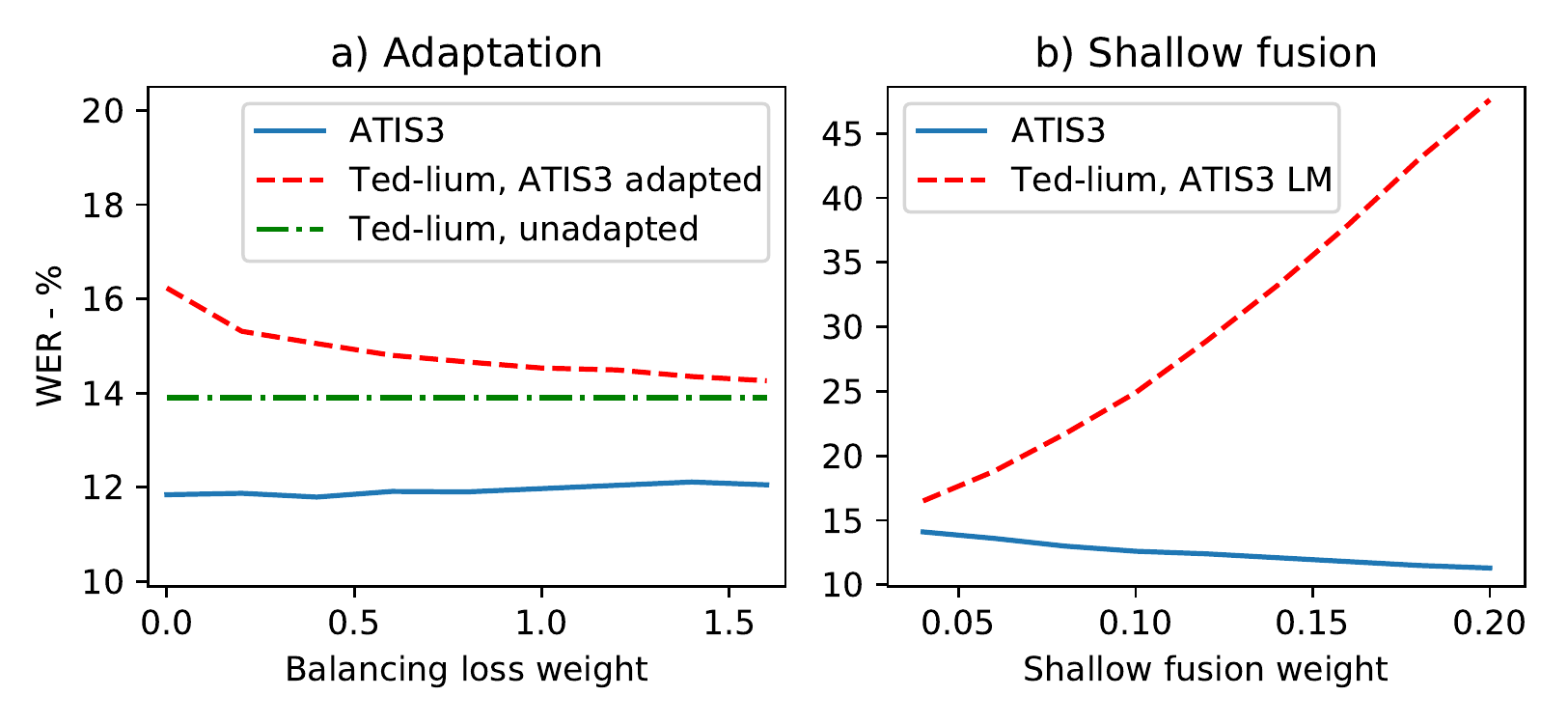}
  \caption{Adaptation experiments with in-domain and out-of-adaptation-domain evaluation. a) The effect of the balancing loss weight to the accuracy of the adapted models. b) Shallow fusion with ATIS3 4-gram, varying the shallow fusion weight.}
  \label{fig:generalization}
\end{figure}

To show the effect of the balancing loss $\ell_b$ in Eq.~\ref{eq:loss}, we used
ATIS3 training data to adapt models with different balancing weights $w_b$. The resulting
models were then tested both with ATIS3 and Ted-lium 3 evaluation sets.
Fig.~\ref{fig:generalization} shows that accuracy on ATIS3 does not
significantly vary with $w_b$. Ted-lium error rates, on the other hand,
reduce as $w_b$ is increased.

Setting $w_b = 0.8$, which was obtained by optimizing the adaptation weights
over the Slurp development set, shows a good compromise between the adapted accuracy
and generalization. With that value, Ted-lium WER degrades from the unadapted baseline
by 5.5\%, but the ATIS3 error reduced by 25\%. A larger balancing weight provides
even better generalization: With value 1.6 Ted-lium results degrade only by 2.6\%
due to ATIS3 adaptation, while ATIS3 WER is still reduced by 24\%.

\subsection{Comparison to shallow fusion}

Shallow fusion \cite{Kannan18} has been shown to perform well with RNN-T models.
It can be trained from a limited amount of text-data and hence applied easily
to customize the ASR system for a target domain. However, it does require some
changes to the decoder.

To study how shallow fusion compares to the presented RNN-T adaption method, we
performed an evaluation with ATIS3 and Ted-lium 3 evaluation sets as in
Section \ref{sec:generalization}, but this time varying the shallow fusion weight.
The shallow fusion LM was a 4-gram model over the word parts, trained from the
ATIS3 adaptation set. The LM was smoothed with absolute discounting.
The results in Fig.~\ref{fig:generalization} show that shallow fusion improves
the in-domain accuracy, as ATIS3 evaluation set WER drops to 11.1\%.
However, this comes at a severe cost to the generalization:
the accuracy on the Ted-lium set degrades
drastically, even with small shallow fusion weights.

Finally we tested a combination of our RNN-T adaptation and shallow fusion.
If the aim is to maximize the accuracy, without
caring about the generalization, this combination provides the best results:
Using ATIS3 adaptation set and the above mentioned 4-gram model provides
a WER of 9.2\% over the ATIS3 evaluation set.

\subsection{Prediction network as a language model}

The improvements obtained by the text-only adaptation of the prediction network $P$
suggest that it does indeed behave a lot like a language model. To analyze this
characteristic further, we ran perplexity experiments with the prediction network and
its LM output $L$ at different stages of the model training, namely the network intialization,
RNN-T training, and adaptation. We used three held-out evaluations sets which match
the data used in those stages:  Oscar corpus subset, LibriSpeech test-clean transcripts, and
ATIS3 adaptation set. Table~\ref{tab:perplexity} summarizes the results.

\begin{table}[t]
  \caption{Word-level perplexities of held-out evaluation text corpora computed with the
  prediction network at different stages of RNN-T training and adaptation.}
  \label{tab:perplexity}
  \centering
  \begin{tabular}{ lccc }
    \toprule
     & \multicolumn{3}{c}{\textbf{Perplexity}} \\
    \textbf{Model} & Oscar & LibriSp & ATIS3 \\
    \midrule
    \#1 Initializing LM & \textbf{123.6} & 286.7  & 238.4 \\
    \#2 RNN-T, old LM output & 151.0 & 292.8  & 276.3 \\
    \#3 RNN-T, new LM output & 179.8 & \textbf{231.4} & 261.9\\
    \#4 ATIS3 adapted RNN-T & 197.9 & 251.9 & \textbf{23.4} \\
    \midrule
    RNN-T, uninitialized $P$ & 1279.2 & 400.5  & 1137.0 \\
    RNN-T, internal LM & 770.5 & 1116.9 & 1055.4 \\
    \bottomrule
  \end{tabular}
\end{table}

Model \#1 is the LM used to initialize $P$ before RNN-T training. It was trained
from the Oscar corpus, so it is natural that it gives the lowest perplexities with the Oscar
evaluation set. After the RNN-T training, but before the estimation of $L$,
we obtain model \#2, which reuses the feed-forward and softmax layers from \#1.
Although the RNN-T training contained LibriSpeech data, we do not see improvements
in the held-out LibriSpeech evaluation set perplexity,
until $L$ is retrained with the speech transcripts (model \#3).
However, the perplexity over the Oscar evaluation set degrades with the replacement of the LM output.
Finally, after the adaptation with the ATIS3 training data (model \#4), we see a
huge improvement in the corresponding evaluation set perplexity.
For comparison, the ATIS3 evaluation set perplexity with a 4-gram model
trained over the adaptation set with word piece segmentation was 39.7.

The conclusion from the perplexity experiments is that the prediction network $P$
and the LM output component $L$ together can perform well as a language model.
During RNN-T training, $P$ learns to better predict
the kind of utterances contained in the training data.
However, to take the full advantage of the LM property of $P$ and $L$,
a new LM output layer needs to be trained after the RNN-T training.
This is because the LM output
layer is analogous to the joint network, and during RNN-T training both the prediction
network and the joint network can change in unison. Nevertheless, the fairly good
perplexities for the model \#2 show that the representations of the trained $P$ are
sufficiently close to the original so that even the output layer of the initializing LM
works. This suggests that the initialization of the prediction network can have
an important role. We verified this by training an RNN-T without initialization,
as suggested by Ghodsi \etal \cite{Ghodsi20}. The corresponding results in
Table~\ref{tab:perplexity} show how the lack of a
proper initialization greatly degrades the LM qualities of the prediction network.
We found that the initialization not only helps in the LM task, but is beneficial
also for ASR accuracy: without the initialization the WER over the Ted-lium 3 evaluation
set degraded from 13.9\% to 15.3\%.

A recent study \cite{Meng21} proposed that the internal language model of an E2E ASR
model should be taken from the softmax output of the joint network. We have adopted
a different view where the prediction network is used with a dedicated LM output layer,
in the same way as it was initialized for the RNN-T training.
For comparison, we tried the "internal LM" \cite{Meng21} approach, but
the perplexities of the LM when taken through the joint network became much worse,
see Table~\ref{tab:perplexity}.

\section{Conclusions}

In this paper we have presented a practical algorithm to perform domain adaptation
of RNN-transducer E2E model with text-only data. The method is fast, requires
no changes to the model inference, and works well even when very little data from
the target domain is available. Compared to popular shallow fusion method,
the presented RNN-T adaptation provides similar accuracy gains, but outperforms
shallow fusion in the generalization capability. This can be contributed to our
fully neural approach where the prediction network is modified under regularizing constraints.

The benefits of the RNN-T adaptation were shown with several evaluation tasks,
using an experiment model which was trained from well-known public speech corpora,
in the interest of experiment reproducibility. We also verified the adaptation gains
with our production model, for which we have used an order of magnitude more training data.
We further showed with LM perplexity experiments that simply by using a separate LM output layer,
the prediction network provides a reasonable performance as a language model.
Our evidence of the improvements obtained from adapting the prediction network lead
us to conclude that the LM interpretation of the prediction network is not only justified,
but also practical.

\section{Acknowledgements}

We thank Business Finland and CSC -- IT Center for Science Ltd. for providing the supercomputer resources used to carry out the experiments in this paper. We also want to thank the Speechly developer community and customers for motivating us to continuously push the boundaries of our work.

\bibliographystyle{IEEEtran}

\bibliography{mybib}

\end{document}